%% file: main.tex
 \documentclass[tablecaption=bottom,wcp]{jmlr} 

\usepackage{booktabs}
\usepackage[load-configurations=version-1]{siunitx} 
\usepackage{natbib}
\usepackage{xcolor}
\usepackage{layouts}

\input{math_commands}
\usepackage[capitalise]{cleveref}

\makeatletter
\newcommand\thefontsize{The current font size is: \f@size pt}
\makeatother

\theorembodyfont{\upshape}
\theoremheaderfont{\scshape}
\theorempostheader{:}
\theoremsep{\newline}

\jmlrproceedings{AABI 2023}{5th Symposium on Advances in Approximate Bayesian Inference, 2023}

\title[Promises and Pitfalls of the Linearized Laplace in Bayesian Optimization]{Promises and Pitfalls of the Linearized Laplace \\ in Bayesian Optimization}

\author{\Name{Agustinus Kristiadi} \Email{akristiadi@vectorinstitute.ai}\\
\addr Vector Institute
\AND
\Name{Alexander Immer} \Email{alexander.immer@inf.ethz.ch}\\
\addr ETH Z\"urich \& MPI-IS T\"ubingen
\AND
\Name{Runa Eschenhagen} \Email{re393@cam.ac.uk}\\
\addr University of Cambridge
\AND
\Name{Vincent Fortuin} \Email{vincent.fortuin@helmholtz-munich.de}\\
\addr Helmholtz AI, Munich
}

\begin{document}

\maketitle

\begin{abstract}
  The linearized-Laplace approximation (LLA) has been shown to be effective and efficient in constructing Bayesian neural networks.
  It is theoretically compelling since it can be seen as a Gaussian process posterior with the mean function given by the neural network's \emph{maximum-a-posteriori} predictive function and the covariance function induced by the empirical neural tangent kernel.
  However, while its efficacy has been studied in large-scale tasks like image classification, it has not been studied in sequential decision-making problems like Bayesian optimization where Gaussian processes---with simple mean functions and kernels such as the radial basis function---are the \emph{de-facto} surrogate models.
  In this work, we study the usefulness of the LLA in Bayesian optimization and highlight its strong performance and flexibility.
  However, we also present some pitfalls that might arise and a potential problem with the LLA when the search space is unbounded.
\end{abstract}



\section{Introduction}
\label{sec:intro}
\input{contents/01_intro}

\section{Linearized Laplace Approximation}
\label{sec:background}
\input{contents/02_background}

\section{The Linearized Laplace in Bayesian Optimization}
\label{sec:bayesopt}
\input{contents/03_bayesopt}

\subsection{Experiment results}
\label{subsec:experiments}
\input{contents/031_experiments}

\subsection{Pitfalls}
\label{subsec:lla_uncertainty}
\input{contents/032_lla_uncertainty}

\section{Conclusion}
\label{sec:conclusion}
\input{contents/04_conclusion}

\acks{
  Resources used in preparing this research were provided, in part, by the Province of Ontario, the Government of Canada through CIFAR, and companies sponsoring the Vector Institute.
  VF was supported by a Branco Weiss Fellowship.
}

{
  \small
  \bibliography{main}
}

\clearpage

\appendix

\section{Experiment Details}
\label{app:details}
\input{contents/91_details}

\section{Additional Experiments}
\label{app:additionals}
\input{contents/92_additionals}

\section{Related Work}
\label{app:related_work}
\input{contents/93_related_work}

\end{document}

%% file: math_commands.tex
\usepackage{amsmath,amssymb,amsfonts,bm,mathtools}

\makeatletter
\def\th@plain{%
  \thm@notefont{}
  \itshape 
}
\def\th@definition{%
  \thm@notefont{}
  \normalfont 
}
\makeatother

\def\1{\bm{1}}

\DeclareMathAlphabet{\mathsfit}{\encodingdefault}{\sfdefault}{m}{sl}
\SetMathAlphabet{\mathsfit}{bold}{\encodingdefault}{\sfdefault}{bx}{n}

\newcommand{\R}{\mathbb{R}}

\DeclareMathOperator*{\argmax}{arg\,max}

\newcommand{\N}{\mathcal{N}}

\renewcommand{\R}{\mathbb{R}}

\newcommand{\D}{\mathcal{D}}

\newcommand{\norm}[1]{\Vert #1 \Vert}

\newcommand{\inv}{{-1}}

\newcommand{\map}{\text{\textnormal{MAP}}}

\newcommand{\X}{\mathcal{X}}
\newcommand{\Y}{\mathcal{Y}}

\newcommand{\lin}{\mathrm{lin}}

%% file: contents/01_intro.tex
Bayesian neural networks (BNNs) have been shown to be useful for predictive uncertainty quantification, aiding tasks such as the detection of out-of-distribution data and adversarial examples \citep{louizos2016structured,louizos2017multiplicative,kristiadi2019cdn}.
Recently, the Laplace approximation \citep[LA,][]{mackay1992bayesian} has emerged to be a compelling practical BNN method and has been successfully deployed in large-scale problems that leverage deep neural nets (NNs), such as image classification and segmentation \citep[etc.]{barbano2022bayesian,miani2022laplacian}, due to its cost-efficiency \citep{daxberger2021laplace} and its \emph{post-hoc} nature.
However, its performance in sequential decision-making problems in small-sample regimes---such as active learning, bandits, and Bayesian optimization---has not been studied extensively.

Nonetheless, the recently proposed linearized version of the LA \citep{immer2021improving} has compelling benefits since it can be viewed as a Gaussian process (GP) model---the \emph{de-facto} standard models used in many sequential problems like Bayesian optimization.
From this perspective, the LLA is a GP with a posterior mean function given by the \emph{maximum-a-posteriori} (MAP) predictive function of the NN and a covariance function given by the NN's empirical neural tangent kernel at the MAP estimate \citep{jacot2018ntk}.
Just like any GP, the LLA can be tuned via its differentiable marginal likelihood using standard deep learning optimizers without validation data \citep{immer2021scalable,immer2022invariance}.
Unlike standard GP models, however, the LLA is much more expressive and accurate due to its NN backbone.
In particular, the LLA inherits the inductive biases of the base NN and enables scalable inference due to its \emph{post-hoc} formulation on top of the MAP-estimated NN.

Given its compelling theoretical benefits and its lack of usage in small-sample problems, it is thus interesting to study the performance of the LLA in this regime.
In this work, we focus on Bayesian optimization and our empirical findings show that the LLA is competitive or even better than the standard GP baseline, especially when the problem requires a strong inductive bias (e.g., search in the space of images).
Nevertheless, the GP view of the LLA also yields a potential pitfall in unbounded domains due to its mean function.
We discuss different ideas on how the LLA can be improved in the future to mitigate this issue.

%% file: contents/02_background.tex
Let \(f: \X \times \Theta \to \Y\) be a neural network (NN) with the input, output, and parameter spaces \(\X \subseteq \R^n\), \(\Y \subseteq \R^k\), and \(\Theta \subseteq \R^d\), respectively.
Given a dataset \(\D := (x_i, y_i)_{i=1}^m\), the standard way to train \(f\) is to find \(\theta_\map \in \Theta\) that maximizes the log-posterior function \(\log p(\theta \mid \D)\).
The \emph{Laplace approximation} (LA) to \(p(\theta \mid \D)\) can then be obtained by computing the Hessian \(H := -\nabla_\theta^2 \log p(\theta \mid \D) \vert_{\theta_\map}\) and letting \(p(\theta \mid \D) \approx \N(\theta \mid \theta_\map, H^\inv)\) \citep{mackay1992evidence}.

The generalized Gauss-Netwon (GGN) matrix is often used to approximate the exact Hessian \citep{martens2020ngd}.
Since the GGN is the exact Hessian of the linearized network \(f^\lin(x; \theta) := f(x; \theta_\map) + J(x) (\theta - \theta_\map)\) where \(J(x) := \nabla_\theta f(x; \theta) \mid_{\theta_\map} \in \R^{k \times d}\) is the Jacobian, it is instructive to use \(f^\lin\) to make predictions under the LA, resulting in the \emph{linearized-Laplace approximation} \citep[LLA,][]{khan2019approximate,immer2021improving}.
While the LLA depends on both the GGN and the Jacobian of the network, it is closely related to second-order optimization\footnote{The derivation of the general LA itself is done by taking a second-order expansion on the loss function.} and thus can leverage recent advances in efficient approximations and computation in that field \citep{dangel2020backpack,osawa2023asdl}.

Due to its weight-space linearity and Gaussianity, the LLA can be seen as a Gaussian process over the function \(f(x)\) directly \citep{immer2021improving,rasmussen2003gaussian}.
That is, it defines a prior \(f \sim \mathcal{GP}(f^\lin, J(x)J(x')^\top)\) over the function, and its posterior on a test point \(x_* \in \X\) is given by the MAP prediction \(f(x_*; \theta_\map)\) with uncertainties given by the standard GP posterior variance under the (empirical, at \(\theta_\map\)) neural tangent kernel \citep[NTK,][]{jacot2018ntk,lee2018deepgp} \(k(x_1, x_2) := J(x_1)J(x_2)^\top\).
In this view, the LLA can be seen as combining the best of both worlds: accurate predictions of a NN and calibrated uncertainties of a GP.
Moreover, many methods have been proposed to improve the efficacy and efficiency of the LLA further \citep[etc]{kristiadi2022frequentist,kristiadi2022refinement,antoran2023sampling,immer2023stochastic,sharma2023incorporating,bergamin2023riemannian}.
Detailed related work is deferred to \cref{app:related_work}.

%% file: contents/03_bayesopt.tex
We evaluate the LLA on three standard benchmark functions: (i) the Branin function on \([-5, 10] \times [0, 15] \subset \R^2\), (ii) the Ackley function on \([-32.768, 32.768]^2 \subset \R^2\), and (iii) the MNIST image generation task minimizing \(f(x) := \norm{x-x_0}\) on \([0, 1]^{784}\) for a fixed \(x_0\) taken from the MNIST training set \citep{verma2022high}.
The base NNs are a three-layer ReLU MLP with \(50\) hidden units each and the LeNet-5 CNN \citep{lecun1998gradient}, and we uniformly sample the initial training set of size \(20\).
Finally, the acquisition function is the popular and widely-used Expected Improvement \citep{jones1998ei}.
See \cref{app:details} for the detailed experimental setup.
Also, refer to \citet{aerni2022laplace} for a comparison of the LLA on sequential decision-making problems, and to the parallel work by \citet{li2023study} for a comprehensive benchmark of BNN-based (including the LLA) Bayesian optimization surrogates against GPs.
The BoTorch implementation of the LLA surrogate is available at \url{https://github.com/wiseodd/laplace-bayesopt}.

%% file: contents/031_experiments.tex
\begin{figure}[t]
  \includegraphics{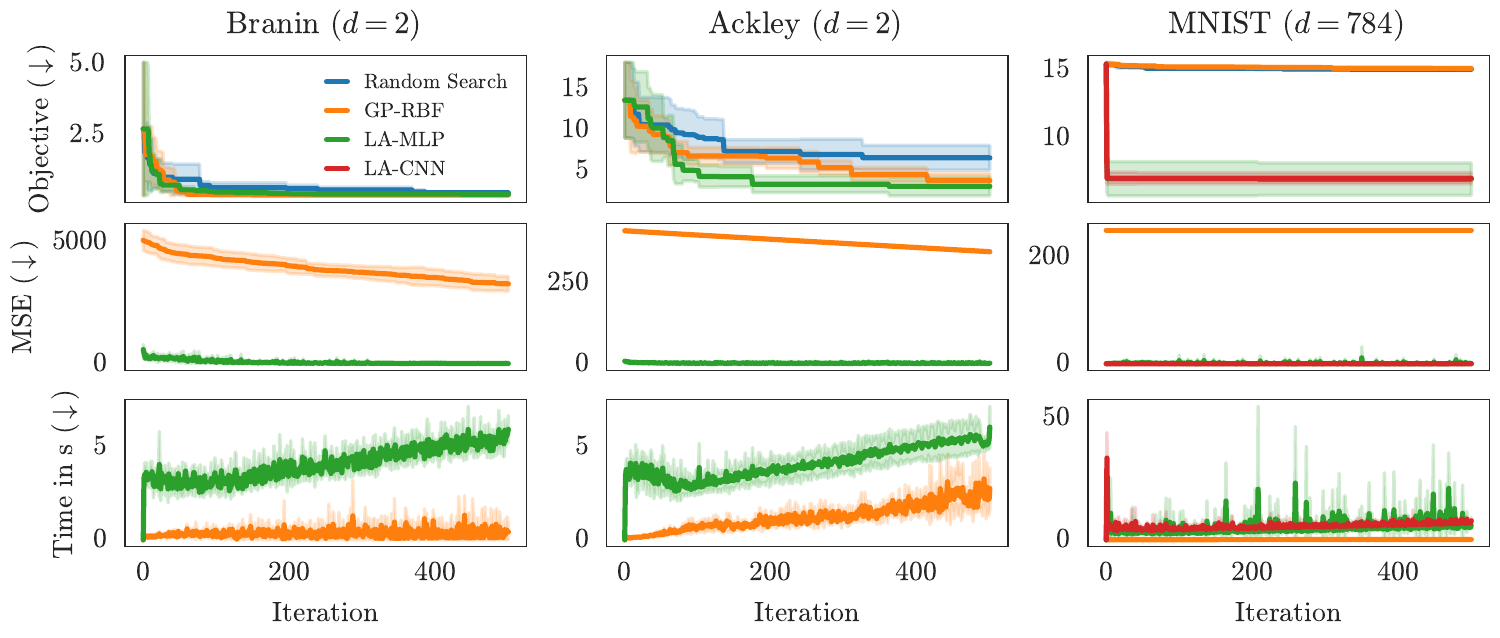}

  \caption{
    Bayesian optimization experiments in terms of objective minimization (\textbf{top}), predictive accuracies of the surrogate models (\textbf{middle}), and the wall-clock time (in seconds) taken to propose a new evaluation point (\textbf{bottom}). The LLA methods outperform the GP on the more complex problems.
  }
  \label{fig:bayesopt_exp}
\end{figure}

The results are shown in \cref{fig:bayesopt_exp}.
In terms of optimization performance, the LLA is competitive or even better than the (tuned) GP baseline on all benchmark functions.
On the high-dimensional MNIST problem, GP-RBF performs similarly (bad) as the random search baseline.
Meanwhile, the LLA finds an \(x\) with small \(f(x)\) in a few iterations.
While the MLP- and CNN-based LLA perform similarly in terms of the mean, the CNN-based LLA is considerably less noisy.
This indicates that LLA-CNN is more reliable that LLA-MLP in finding the minimizers of the objective function.
All in all, the better performance of the LLA can be contributed to the strong predictive performance (in terms of test mean-squared error\footnote{The test set is uniformly sampled from the search space.}) of the base NNs.

In terms of computation cost, the LLA induces some overhead compared to the (exact) GP-RBF.
This is because, at each iteration, the acquisition function needs to be optimized by backpropagating through the NN and its Jacobian.
Nevertheless, we found that the LLA is more memory-efficient than the exact GP due to the fact that it can leverage mini-batching.
Meanwhile, the GP can quickly yield an out-of-memory error (after around \(800\) iterations).
While both GPs and the LLA can leverage sparse posterior approximations \citep{titsias2009variational}, unlike standard GPs, the LLA has the advantage that it is \emph{both} parametric and non-parametric (\cref{sec:background}) and thus one can pick a cheaper inference procedure (i.e., in weight space or function space) depending the amount of data at hand and the size of the network \citep{immer2023stochastic}.

\begin{figure}[t]
  \includegraphics{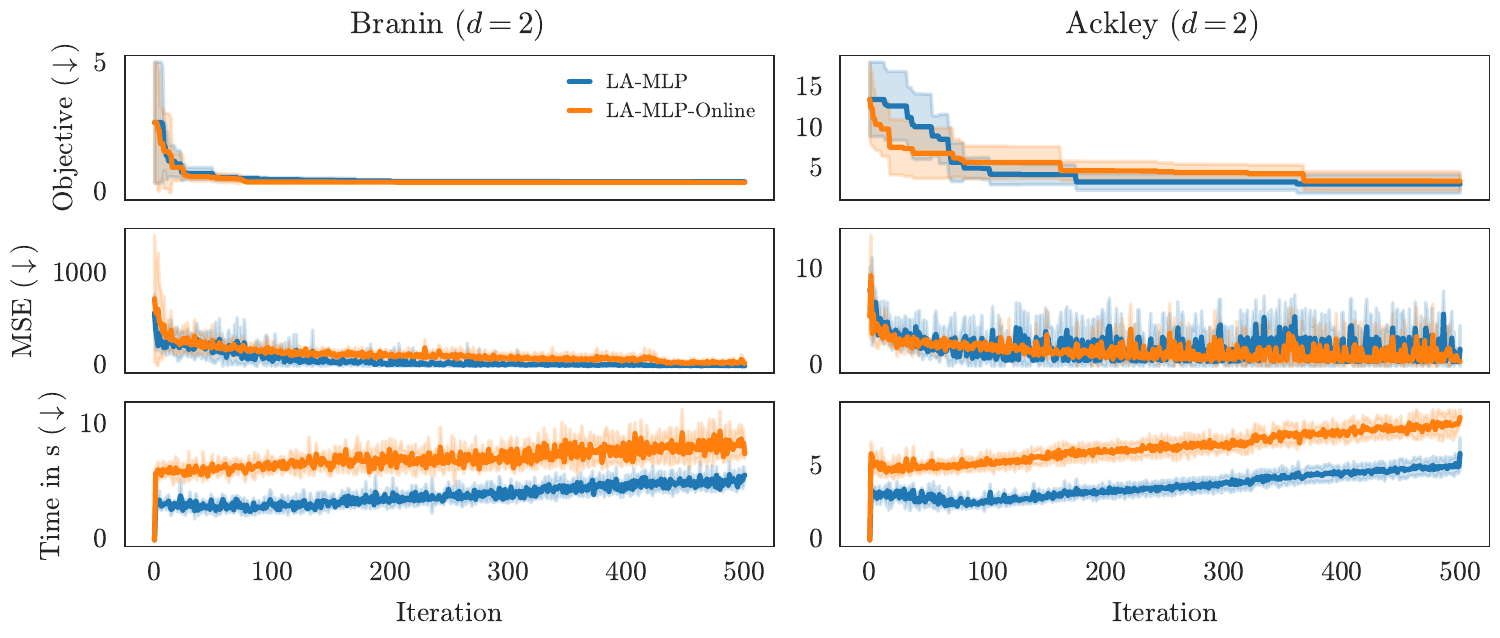}

  \caption{
    \emph{Post-hoc} vs.\ online LLA in Bayesian optimization, in terms of objective minimization (\textbf{top}), predictive accuracies of the surrogate models (\textbf{middle}), and the wall-clock time (in seconds) taken to propose a new evaluation point (\textbf{bottom}). The \emph{post-hoc} LLA tends to perform at least on par with the online one while being computationally cheaper.
  }
  \label{app:fig:bayesopt_laplaces}
\end{figure}

Finally, we compare the two popular methods for tuning LLA: \emph{online} \citep{immer2021scalable} and \emph{post hoc} \citep{kristiadi2020being,kristiadi2021lula,eschenhagen2021mixtures}.
We found that \emph{post-hoc} LLA performs similarly to the online LLA while being cheaper since the marginal likelihood tuning is only being done once after the MAP training (see \cref{app:fig:bayesopt_laplaces}).
This is encouraging because the strong performance of the LLA thus can be obtained at a low cost.
Note that, this finding agrees with the conclusion of \citet{aerni2022laplace}.

In \cref{app:additionals}, we further show empirical evidence regarding the capabilities of the LLA in small-sample regimes.
These results further reaffirm the effectiveness of the LLA in Bayesian optimization, and possibly in other sequential problems.

%% file: contents/032_lla_uncertainty.tex
\begin{figure}[t]
  \includegraphics{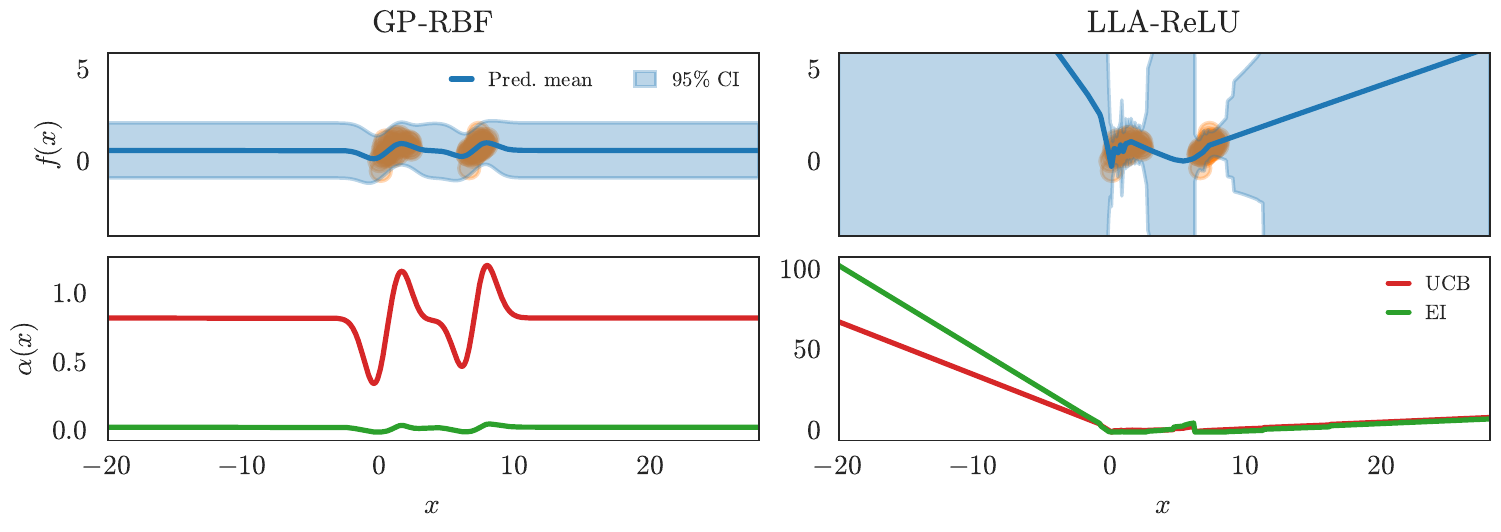}

  \caption{
    As a surrogate function, LLA with ReLU activations is (pathologically) always increasing/decreasing far away from the data.
    This impacts popular acquisition functions like UCB and EI and thus the exploration-exploitation behavior.
  }
  \label{fig:gp_vs_lla}
\end{figure}

We have seen empirically that the LLA can be useful in sequential problems with \emph{bounded} domains.
However, the function-space interpretation of the LLA posterior---MAP-estimated NN as its mean function and the corresponding empirical (i.e.\ finite-width) NTK kernel as its covariance function---can be a double-edged sword in general.
Specifically, \citet{hein2019relu} have shown that ReLU networks, the most popular NN architecture, perform pathologically in the extrapolation regime outside of the data in the sense that \(f(\,\cdot\,; \theta_\map)\) is a non-constant linear function, i.e., always increasing or decreasing.
Thus, when a ReLU NN is used in the LLA, the LLA will inherit this behavior.
As a surrogate function, this ReLU LLA will then behave pathologically under commonly-used acquisition functions \(\alpha\):
The acquisition function will always pick \(x \in \X\) that maximizes \(\alpha\) under \(f(x; \theta_\map)\) and its uncertainty---but when \(f(x; \theta_\map)\) is always increasing (or decreasing) outside the data, \(\alpha\) will always yield points that are far away from the current data (see \cref{fig:gp_vs_lla}).

While \citet{kristiadi2020being,kristiadi2021rgpr} have shown that the LLA's uncertainty can be sufficient in counterbalancing the linear growth of the mean function in classification, this mitigation is absent in regression---the standard problem formulation in Bayesian optimization.
For instance, let
\begin{equation}
  \alpha(x) := f(x; \theta_\map) + \beta \sigma(x)
\end{equation}
with \(\beta > 0\) be the UCB \citep{garivier2011upper} under the LLA posterior, where \(\sigma^2(x)\) is the GP posterior variance on \(x\) under the empirical NTK kernel.
Then, since outside of the data region, \(f(x; \theta_\map)\) and \(\sigma(x)\) is always increasing/decreasing linearly in \(x\) \citep{kristiadi2020being}, it is easy to see that \(\alpha(x)\) is also increasing/decreasing linearly in \(x\).
Since the goal is to obtain a next point \(x_{t+1} = \argmax_x \alpha(x)\) that maximizes/minimizes \(\alpha\), the proposed \(x_{t+1}\) will always be far away from the current data (infinitely so when \(\X\) is an unbounded space).\footnote{A similar argument also applies to the expected improvement acquisition function \citep{jones1998ei}.}

\begin{figure}[t]
  \includegraphics{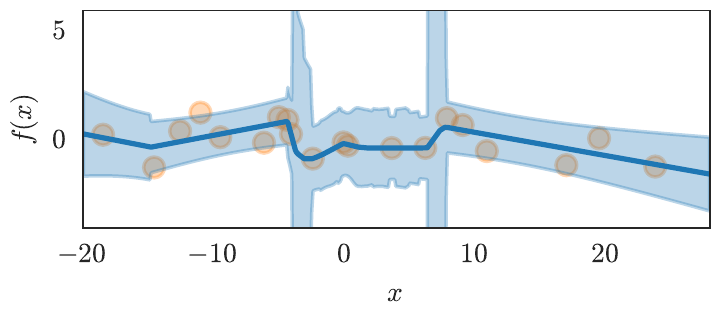}
  \includegraphics{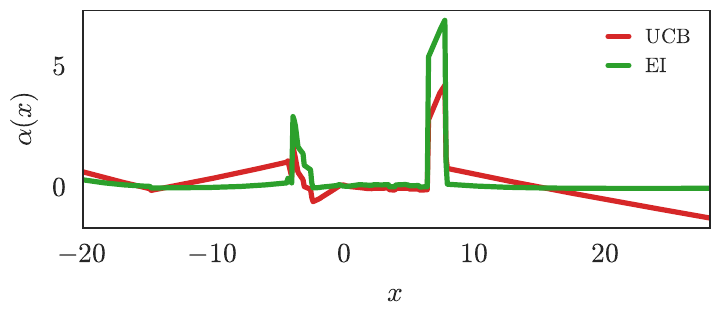}

  \caption{
    By (i) specifiying bounds on \(x\), (ii) initialize the training data uniformly at random, one can mitigate the pathology to some degree.
  }
  \label{fig:lla_mitigation}
\end{figure}

While solving this issue is beyond the scope of the present work, one can follow the following recommendation to mitigate the aforementioned pathology, to some degree (\cref{fig:lla_mitigation}):
(i) specify the upper and lower bounds on the possible values of \(x\), and (ii) use random points, sampled uniformly from inside the bounds, to initialize the training data for the LLA.

Beyond the above ``hotfix'', possible solutions are (i) better architectural design, e.g., using more suitable activation functions \citep{meronen2021periodic}, and (ii) designing an acquisition function that takes the behavior induced by the MAP-estimated network into account.
For instance, one option for the latter could be to avoid the asymptotic regions of the ReLU network altogether.
While this restricts the exploration to be regions near the already gathered data, similar to proximal or trust-region methods \citep{schulman2015trust, schulman2017proximal}, it might be worth paying this price to harness the capabilities of neural networks, like strong inductive bias and good predictive accuracy.

%% file: contents/04_conclusion.tex
We have empirically validated the effectiveness of the linearized Laplace in Bayesian optimization problems:
It can outperform standard Gaussian process baselines, especially in tasks where strong inductive biases are beneficial, e.g., when the search space consists of images.
The linearized Laplace provides flexibility to the practitioners in incorporating the inductive bias due to its Gaussian-process interpretation: its posterior mean is given by the neural network's MAP predictive function and its posterior mean is induced by the empirical neural tangent kernel of the network.
The former implies that the predictive mean of the linearized Laplace directly encodes the inductive bias of the base NN.

On the other hand, the aforementioned properties of the linearized Laplace yield a pitfall in Bayesian optimization.
Namely, the MAP predictive functions of ReLU networks are known to be pathological: they are always increasing or decreasing outside of the data.
Moreover, the induced neural tangent kernel is non-stationary, unlike the commonly-used kernel in Bayesian optimization.
They can thus yield pathological behavior in unbounded domains when used in conjunction with standard acquisition functions.
The investigation of this issue is an interesting direction for future work.

%% file: contents/91_details.tex
In the following, we describe the experimental setup we used to obtain the Bayesian optimization results presented in the main text.

\paragraph*{Models} 
For the MLP model, the architecture is \(n-50-50-1\). 
Meanwhile, for the CNN model, the architecture is the LeNet-5 \citep{lecun1998gradient} where the two last dense hidden layers are of size \(50\) and \(20\).

\paragraph*{Training}
For all methods (including the baselines), we initialize the starting dataset by uniformly sampling the search space and obtain the target by evaluating the true objective values on those samples.
For both models, we train them for \(1000\) epochs with full batch ADAM and cosine-annealed learning rate scheduler.
The starting learning rate and weight decay pairs are (\num{1e-1}, \num{1e-3}) and (\num{1e-3}, \num{5e-4}) for the MLP and CNN models, respectively.
For the LLA, the marginal likelihood optimization is done for \(10\) iterations.
For the online LLA, this optimization is done every \(50\) epochs of the ``outer'' ADAM optimization loop.
For the other hyperparameters for the marginal likelihood optimization, we follow \citet{immer2021scalable}.

\paragraph*{Baselines}
The random search baseline is done by uniformly sampling from the search domain.
Meanwhile, for the GP-RBF baseline, we use the standard setup provided by \texttt{BoTorch} \citep{Balandat2020BoTorchAF}.
In particular, it is tuned by maximizing its marginal likelihood.

%% file: contents/92_additionals.tex
\begin{figure}[t]
  \includegraphics{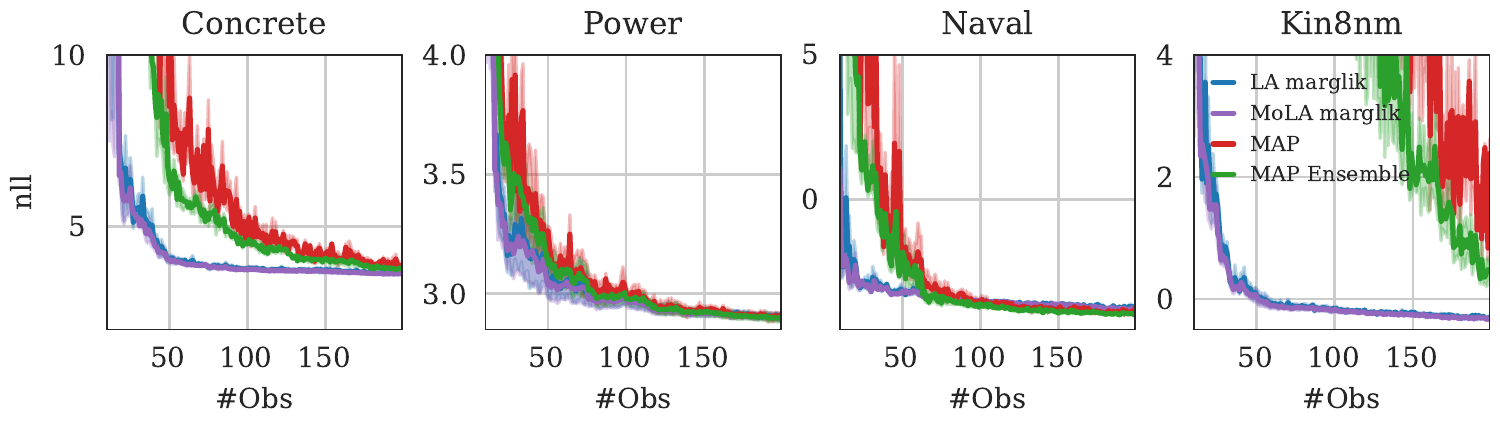}
  \caption{
    Test negative log-likelihood (nll) for UCI regression datasets with MAP prediction (not Bayesian) of MAP optimized models and model trained with marginal likelihood optimization (``marglik'') during training as in \citep{immer2021scalable}. 
    All methods were trained on \(1-200\) observation points, where in each turn, one observation point is added.
    ``MoLA'' indicates a mixture of Laplace approximations with the same number of components as the deep ensemble. The online marginal-likelihood optimization methods outperform simple MAP with fixed prior precision.
  }
  \label{fig:uci_map}
  
  \includegraphics{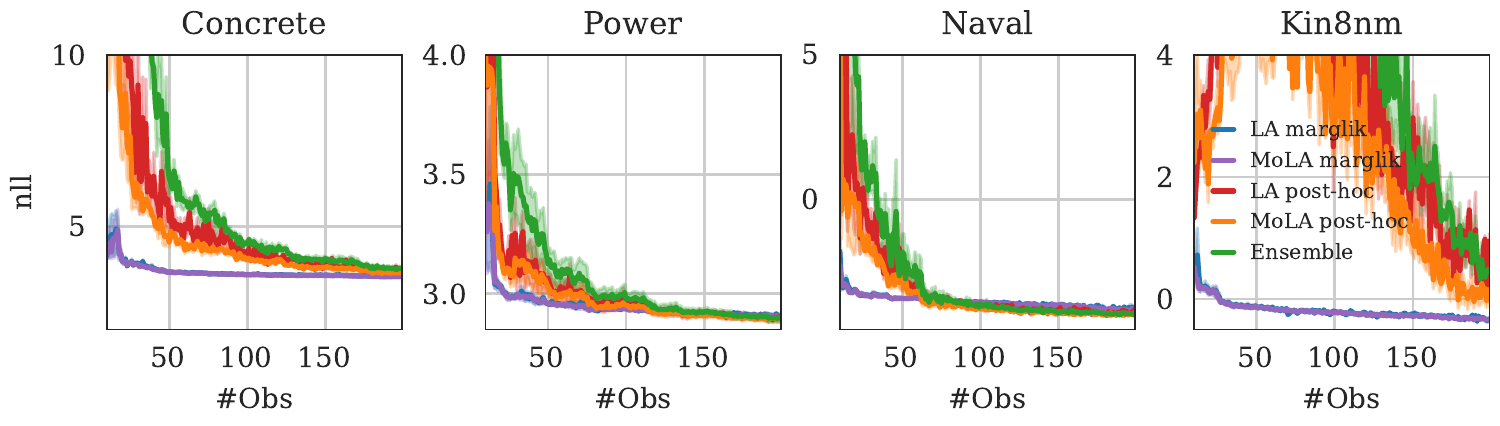}
  \caption{
    Test negative log-likelihood (nll) on the same four UCI regression data sets but instead with the Bayesian posterior predictive of the Laplace approximation (i.e.\ via Bayesian model averaging) in comparison to the same deep ensemble as in the figure above. The online marginal-likelihood optimization methods again clearly outperform the others. ``\emph{post-hoc}'' is a Laplace approximation trained with a fixed prior precision and optimizes it after training.
    Interestingly, the Bayesian predictive helps the unimodal MAP to improve over the deep ensemble and MoLA does so even more.
    Overall, using the Bayesian predictive tends to improve the performance for all models.
  }
  \label{fig:uci_bayes}
\end{figure}

\subsection{Small-Data Regression}
\label{subsec:small_data}

\paragraph*{Results}
In \cref{fig:uci_map} and \cref{fig:uci_bayes}, we compare the Laplace approximation methods to a deep ensemble of five neural networks on four UCI regression datasets.
The models are trained on small increasing subsets from one to \(200\) data points and evaluated on a held-out validation set consisting of the held-out data points. 
The mixture of Laplace (MoLA) variants are trained with the same number of components as the deep ensemble.
We show results for \emph{post-hoc} Laplace approximation, which is trained with a fixed prior precision and optimizes it after training~\citep{mackay1992bayesian,eschenhagen2021mixtures}, 
and \emph{marglik} Laplace approximation, which is trained with Laplace marginal likelihood optimization during training~\citep{immer2021scalable}.

The results show that the online marginal likelihood optimization performs best early during training with the mixture and single Laplace performing similarly well. 
Further, the methods clearly improve when using the Bayesian posterior predictive in \cref{fig:uci_bayes} compared to the MAP prediction in \cref{fig:uci_map}.
This is clearly visible by comparing \emph{post-hoc} MoLA to the Ensemble method \citep{lakshminarayanan2017simple}, which is the same model but predicts with the MAP of the individual models instead of their Laplace approximation~\citep{eschenhagen2021mixtures}.
Even \emph{post-hoc} Laplace can outperform the Ensemble, despite only using a single model.

\paragraph*{Setup}
All models are MLPs with fifty hidden units and ReLU activation.
We train the methods on one to \(200\) training points, in each turn adding a single data point, and repeat this for three runs.
The test (predictive) log-likelihood is reported over three runs with standard error.
All models are trained for \(1000\) steps using Adam~\citep{kingma2014adam} with a learning rate of \num{1e-3} decayed to \num{1e-5}.
The online marginal likelihood variant optimizes every \(50\) steps for \(50\) hyperparameter steps and uses early stopping on the marginal likelihood value~\citep{immer2021scalable}.
The \emph{post-hoc} variant optimizes the hyperparameters after training for \(1000\) steps and uses the last-layer posterior predictive instead of the full-network predictive since it becomes less stable when applied after training without optimizing the prior precision with either cross-validation or the marginal likelihood during training.

\begin{table}
  \centering
  \begin{tabular*}{0.75\linewidth}{@{\extracolsep{\fill}}lcc}
    \toprule
    \textbf{Method}     & \textbf{Normalized KL} $\downarrow$     & \textbf{Test Accuracy} $\uparrow$ \\
    \midrule
    Uniform  & 1.000 $\pm$ 0.007 & 0.4735  \\
    MLP      & 0.391 $\pm$ 0.171 & 0.9159  \\
    Ensemble & 0.381 $\pm$ 0.163 & 0.9161  \\
    LA \textit{post-hoc}    & 0.643 $\pm$ 0.064 & 0.9147  \\
    LA \textit{marglik}    & 0.567 $\pm$ 0.064 & 0.9157  \\
    \bottomrule
  \end{tabular*}
  \caption{Results on the \texttt{CLASSIFICATION\_2D\_TEST} sweep of the Neural Testbed. Surprisingly, both Laplace approximation variations are worse than the simple baselines in terms of the KL-metric proposed by \citep{wen2021predictions}.}
  \label{tab:neural_testbed}
\end{table}

\subsection{Evaluating Joint Predictions on the Neural Testbed}
Traditionally, the Bayesian deep learning literature focuses on marginal predictions on individual data points (c.f. \cref{app:related_work}).
However, \citet{wen2021predictions} argue that \emph{joint predictive distributions}, i.e. predicting a set of labels from a set of inputs, which can capture correlations between predictions on different data points, is much more important for downstream task performance, e.g. in sequential decision making.
To facilitate the evaluation of the joint predictive distribution of agents, \citet{osband2022neural} propose a simple benchmark of randomly generated classification problems, called the \emph{Neural Testbed}.
While it only consists of small-scale and artificial tasks, the goal is to have a ``sanity check'' for uncertainty quantification methods in deep learning, to judge their potential on more realistic problems and help guide future research.
The authors also show empirically that the performance of the joint predictive is correlated with regret in bandit problems \citep{gittins1979bandit}, whereas the marginal predictive performance is not.

Recall that the linearized Laplace is a GP: 
It naturally gives rise to a joint predictive distribution over input points.
However, Laplace approximation-based methods have not been evaluated on this benchmark yet.
Here, we include preliminary results on the \texttt{CLASSIFICATION\_2D\_TEST} subset of the benchmark\footnote{Available at \url{https://github.com/deepmind/neural_testbed}.}.

\paragraph*{Setup}
As in the experiments in \cref{subsec:small_data}, all models are MLPs with fifty hidden units and ReLU activations.
We optimize for \(100\) epochs using Adam with a learning rate of \num{1e-3}.
We train on \texttt{CLASSIFICATION\_2D\_TEST}, a set of seven \(2\)D binary classification problems.
Each method is evaluated on a test set by calculating an approximation to the expected Kullback-Leibler (KL) divergence between the predictive distribution under the method and the true ground-truth data generating process.
We normalize the KL divergence to \(1.0\) for our reference agent which predicts uniform class probabilities.
See \citet{osband2022neural} for more details.

For the \emph{post-hoc} tuning of the prior precision we optimize the marginal likelihood once after training and use a last-layer Laplace approximation with full covariance.
For the online tuning of the prior precision via optimizing the Laplace marginal likelihood, we use a burn-in of \(20\) steps and then take \(50\) optimization steps every five training steps.
Here, we use a Laplace approximation over all weights with a Kronecker factored covariance and a layer-wise prior precision.
For both variations of the method, we approximate the predictive using the probit approximation \citep{spiegelhalter1990sequential,mackay1992evidence}.
In addition, we tried different predictive approximations, but got comparable results.

\paragraph*{Results}
In \cref{tab:neural_testbed}, we can see results for a regular MLP, a deep ensemble, and the Laplace approximation.
Surprisingly, both variations of the LA perform significantly worse in terms of KL divergence compared to the MLP and the deep ensemble, while the test accuracy is comparable.
Online marginal likelihood optimization performs better than \emph{post-hoc}, but since both are so much worse than the simple baselines, the significance of this difference is unclear.
Since we would expect the Laplace approximation to do at least as well as the MLP, we think further investigating the discrepancy in performance is a important step towards applying the Laplace approximation to sequential decision making problems.
Also, the results demonstrate the potential pitfalls of expecting Bayesian deep learning techniques to work out-of-the-box in new settings.

%% file: contents/93_related_work.tex
\newcommand{\addcite}[1]{\textcolor{red}{[CITE: #1]}}

\paragraph{Bayesian neural networks}

Bayesian neural networks promise to combine the expressivity of neural networks with the principled statistical properties of Bayesian inference \citep{mackay1992practical, neal1993bayesian}.
However, since their inception, performing approximate inference in these complex models has remained a lingering challenge \citep{jospin2022hands}.
Since exact inference is intractable, approximate inference techniques follow a natural tradeoff between quality and computational cost, from cheap local approximations like Laplace inference \citep{laplace1774memoires, mackay1992practical, khan2019approximate, daxberger2021laplace}, stochastic weight averaging \citep{izmailov2018averaging, maddox2019simple}, and dropout \citep{gal2016dropout, kingma2015variational}, via variational approximations with different levels of complexity \citep[e.g.,][]{graves2011vi, blundell2015bbb, louizos2016structured, khan2018vogn, osawa2019practical}, across ensemble-based methods \citep{lakshminarayanan2017simple, wang2019function, wilson2020bayesian, ciosek2020conservative, he2020bayesian, d2021stein, d2021repulsive, eschenhagen2021mixtures}, up to the very expensive but asymptotically correct Markov Chain Monte Carlo (MCMC) approaches \citep[e.g.,][]{neal1993bayesian, neal2011mcmc, welling2011bayesian, garriga2021exact, izmailov2021hmc}.
Apart from the challenges relating to approximate inference, recent work has also studied the question of how to choose appropriate priors for BNNs \citep[e.g.,][and references therein]{fortuin2021bnnpriors, fortuin2022bayesian, nabarro2022data, sharma2023incorporating, fortuin2022priors} and how to effectively perform model selection in this framework \citep[e.g.,][]{immer2021scalable,immer2022probing, immer2022invariance, rothfuss2021pacoh, rothfuss2022pac, van2022learning, schwobel2022last}.
In our work, we particularly draw on the recent advances in Laplace inference \citep{daxberger2021laplace}, particularly linearized Laplace \citep{mackay1992evidence,immer2021improving}, and its associated marginal-likelihood-based model selection capabilities \citep{immer2021scalable}. However, we apply these methods to the sequential learning setting, especially Bayesian optimization, which to the best of our knowledge has not been studied before.

\paragraph{Bayesian optimization}

While Bayesian optimization (BO) has been intensely studied \citep[][and references therein]{shahriari2015taking, garnett2023bayesian}, the \emph{de-facto} standard models for this purpose have been Gaussian processes (GPs) \citep{rasmussen2003gaussian, Balandat2020BoTorchAF, gardner2018gpytorch}.
Indeed, existing work using BNNs as surrogate models for BO is scarce \citep{snoek2015scalable, springenberg2016bayesian, kim2021scalable, rothfuss2021pacoh} and all these approaches use either Bayesian linear regression \citep{snoek2015scalable}, MCMC \citep{springenberg2016bayesian}, or ensemble-based inference \citep{kim2021scalable, rothfuss2021pacoh}.
To the best of our knowledge, we are the first to propose Laplace inference for this problem and highlight its performance benefits, low computational cost, and model selection capabilities.